\pdfoutput=1
\documentclass[11pt]{article}

\usepackage{EMNLP2022}

\usepackage{times}
\usepackage{latexsym}
\usepackage{tabularx}
\usepackage{makecell}
\usepackage{booktabs}

\usepackage{graphics}

\usepackage[T1]{fontenc}
\usepackage[utf8]{inputenc}

\usepackage{microtype}

\usepackage{graphicx}
\usepackage{multirow, hhline, graphicx, subcaption, helvet}
\mathchardef\mhyphen="2D

\newenvironment{enumeratesquish}{\begin{list}{\addtocounter{enumi}{1}\labelenumi}{\setlength{\itemsep}{0em}\setlength{\labelwidth}{0.5em}\setlength{\leftmargin}{\labelwidth}\addtolength{\leftmargin}{\labelsep}}}{\end{list}\setcounter{enumi}{0}}

\usepackage{xspace}
\newcommand{\base}{\textsc{base}\xspace}
\newcommand{\srl}{\textsc{srl}\xspace}
\newcommand{\cefr}{\textsc{cefr}\xspace}
\newcommand{\wiki}{\textsc{wiki}\xspace}

\newcommand{\commongen}{\textsc{commongen}\xspace}
\newcommand{\vocabulary}{\textsc{vocabulary}\xspace}

\makeatletter
\newcommand{\printfnsymbol}[1]{%
  \textsuperscript{\@fnsymbol{#1}}%
}

\title{Controlled Language Generation for Language Learning Items}

\author{Kevin Stowe\thanks{~~Equal Contribution.} , 
  Debanjan Ghosh\printfnsymbol{1}, 
  Mengxuan Zhao\\ 
  Educational Testing Service \\
  {\tt \{kstowe, dghosh\}@ets.org}, {\tt mzhao@etscanada.ca}
  }

\begin{document}
\maketitle
\begin{abstract}

    This work aims to employ natural language generation (NLG) to rapidly generate items for English language learning applications: this requires both language models capable of generating fluent, high-quality English, and to control the output of the generation to match the requirements of the relevant items. We experiment with deep pretrained models for this task, developing novel methods for controlling items for factors relevant in language learning: diverse sentences for different proficiency levels and argument structure to test grammar. Human evaluation demonstrates high grammatically scores for all models (3.4 and above out of 4), and higher length (24\%) and complexity (9\%) over the baseline for the advanced proficiency model. Our results show that we can achieve strong performance while adding additional control to ensure diverse, tailored content for individual users.\footnote{Code and datasets made available at \url{https://github.com/EducationalTestingService/concept-control-gen}}

\end{abstract}

\section{Introduction} 
\label{section:intro}

% Two specific things
%  Personalization (ie. for difficulty)
%  Grammar (ie. SRL)

Recent advancement of transformer \cite{DBLP:conf/nips/VaswaniSPUJGKP17} based pre-trained language models (LM) \cite{lewis-etal-2020-bart, brown2020language,raffel2020exploring} have resulted in unprecedented success in generating large amounts of fluent English text. %add more  
%Natural Language Generation (NLG) is a powerful tool usable for generating large amounts of fluent English text 
One possible area where text generation can be applied is item generation for English language learning applications (LLAs). LLAs are  popular apps used by millions of people all over the world.\footnote{\url{https://www.businessofapps.com/data/language-learning-app-market/}} These apps often include multiple choice items for vocabulary tests, flashcards, grammar lessons, and more.
% later: In this work, we are focused on generating sentence items for testing user vocabulary, both in terms of comprehension as well as production. 
Typically, such items are created manually \cite{educational2010toefl} or curated from crowd-sourced sentence database, e.g., Tatoeba \cite{settles2020machine}.\footnote{https://tatoeba.org/en/} On the contrary, our goal is to make this process scalable by employing LMs, enabling developers of LLAs to be able to implement a much broader array of learning items quickly and efficiently. 
%\Debanjanin{what do you mean by minimal additional manual effort.}

%    There is a wide body of work relating to controlling language generation: our work is novel in its application to the generation of learning assessment items. We provide a framework for the creation of a variety of possible items, and implement methods for specifically controlling not only the type of item generated but also relevant stylistic factors: the argument structure and relative difficulty of the sentence, as well as the relevant vocabulary words contained therein.

\begin{figure}[t]
    \centering
    \includegraphics[width=.5\textwidth]{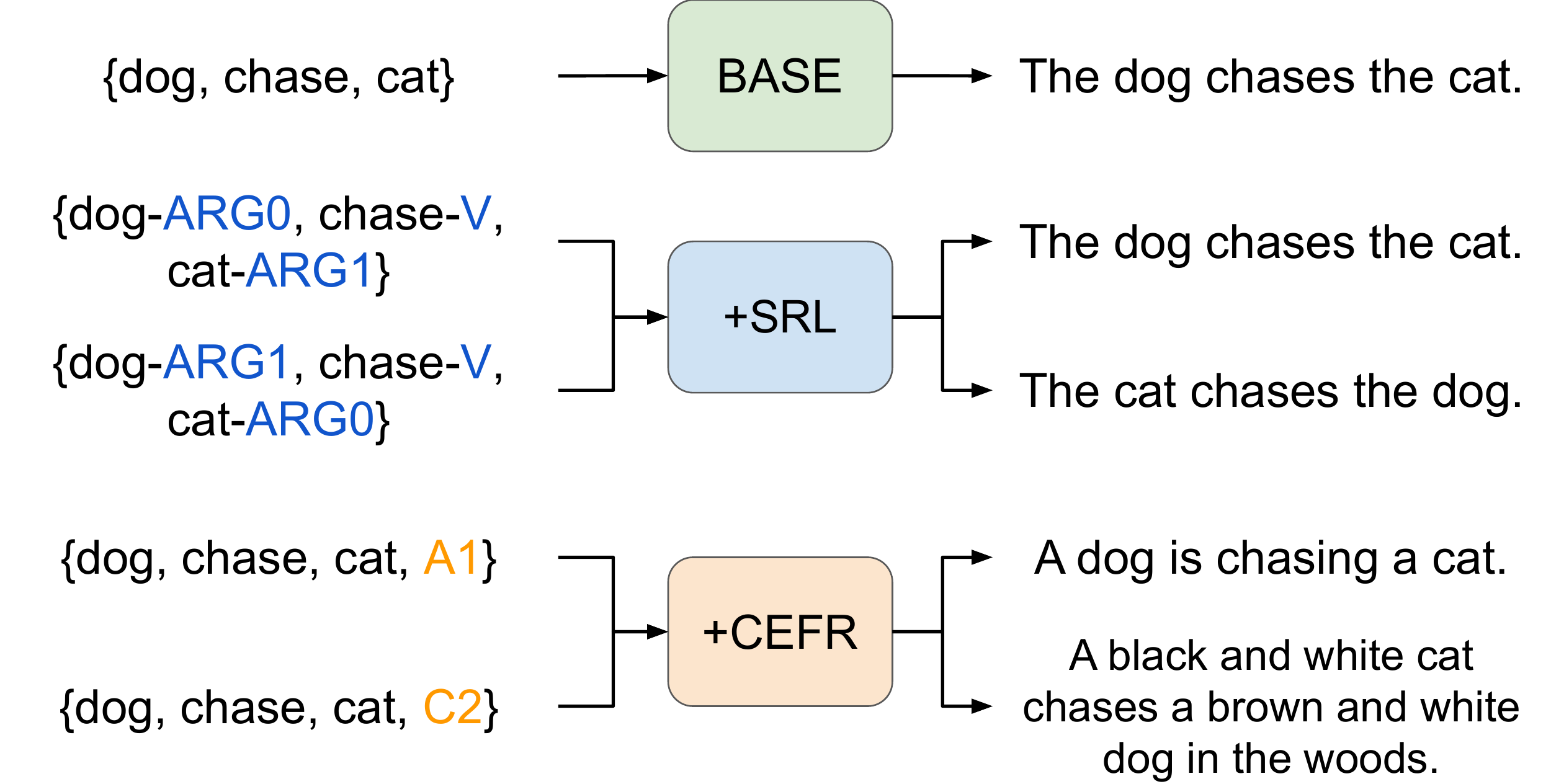}
    \caption{Controlling the concept2seq generation process, using semantic role labels and CEFR levels.}
    \label{fig:introduction}
\end{figure}

This is accomplished by using a  \emph{concept2seq} framework: a sequence-to-sequence architecture in which, given a set of relevant \emph{concepts}, we aim to generate sentences that minimally contain those concepts.\footnote{Concepts are lemmatized tokens in text extracted by using the ConceptNet knowledge base \cite{speer2017conceptnet}.} There is a wide body of work relating to this framework \cite{lin-2020,carlsson-etal-2022-fine} to generate sentences: we experiment with a number of adaptations that are applicable to LLAs. First, the importance of providing diverse content based on student/user needs and different skill sets is well established in learning sciences \cite{morgan2014maximizing,clarke2003changing}. We generate sentences according to different skill sets by conditioning on the Common European Framework of Reference for Languages (CEFR) levels in English.\footnote{CEFR is an international standard for measuring user' ability within a language.} We use a document level CEFR predictor \cite{montgomerie-2022} to predict the CEFR levels in the training data and in turn use the level in a controlled generation framework based on BART \cite{lewis-etal-2020-bart} to generate different CEFR level-specific sentence (Section \ref{sec:cefr}). Second, to test the grammatical proficiency of the users we attach syntactic and semantic information in the form of semantic role labels (SRL) to the concept inputs for controlled generation where we can specify the semantic roles (e.g., ARG0) in the generation (Section \ref{sec:srl}). Such generations can be used in LLAs to ask grammar questions and further expand the diversity of items.  

Consider the examples in Figure \ref{fig:introduction}. The \base model generates a  sentence using the concepts ``dog'', ``chase'', and ``cat''. The \cefr model demonstrates two different generations for two skill levels: a simple short sentence (e.g. ``a cat is chasing a dog'') for the A1 beginner level and a long complex sentence (e.g., ``a black and white \dots in the woods) for the C2 proficiency level (Section \ref{sec:cefr}). Likewise, the \srl model presents examples where we conditioned the generations on specific semantic roles (Section \ref{sec:srl}).
%cat and dog explanation is space is available 

%adding information about sentence difficulty to control for different levels of items \cite{north2014cefr}, (b) adding additional syntactic and semantic information in the form of dependency parse and semantic role labels to the concept inputs, and (c) additional pretraining with semi-supervised concept2seq data. Consider the examples in Table \ref{table:one} - 

%exploring multiple encoding streams to better capture the relationships between the concepts and their argument structure.

We evaluate our models using automatic metrics relevant to our goals (perplexity, concept coverage in the generated sentences, length, and lexical diversity), as well as utilizing Amazon Mechanical Turk (MTurk) to get human judgments of important factors: grammaticality, complexity, and semantic plausibility. The CEFR model generates less complex, shorter sentences than the baseline when targeting the A1 level (complexity score of 2.45, average length of 11.6) compared to the C2 level (complexity score of 2.73, average length of 15.3 words). The SRL model generates the targeted words in the correct argument slot significantly more than the baseline (improving from 6\% to 32\% based on the targeted role). All models are within 3\% of the baseline in terms of grammaticality and semantic plausibility, indicating that we can effectively generate sentences from concepts while adding additional control.
\section{Data} 
\label{section:data}

We employ two datasets for concept2seq generation. Each instance in the datasets is a set of concepts paired with a sentence that contains those concepts. First, we use the \commongen data which is based on existing caption corpora \cite{lin-2020}. From this dataset we use 71,408 concept/sentence pairs. Although \commongen is used in related concept2seq generation \cite{lin-2020}, since it is based on image captions, many samples are phrases (not sentences) and less diverse. Thus, we also collect another dataset based on fourteen relevant vocabulary items for language learning that belong to different CEFR levels.\footnote{This vocabulary list and their word derivatives was recommended by the learning scientists of the LLA.}  Sentences are collected from diverse sources such as the ROCStories \cite{mostafazadeh2016corpus}, Tatoeba sentence database, and the Google book corpus.\footnote{\url{https://www.english-corpora.org/googlebooks/}} We first retrieved sentences containing the vocabulary items from these different sources and then extracted the concepts using the ConceptNet knowledge-base by employing \newcite{becker-etal-2021-coco}. We extract noun, verbs, and adjective tokens as concepts and keep only those sentences containing 2-5 concepts to keep consistency with \commongen. This dataset is denoted as the \vocabulary dataset and consists of an additional 218,997 concept/sentence pairs. In total, the \commongen and \vocabulary gives us a dataset of 290,399 pairs of concepts and sentences. 

%where we used 80\% of the dataset for training, 10\% for validation and remaining 
    
To evaluate our generation models, we create two test sets to evaluate two different scenarios. Our goal is to evaluate concept sets that occur frequently in the training data, as well those that occur rarely. In order to build these two types of test sets, we first generate the frequency counts of each concept over the entire dataset. We then calculate the frequency of a given concept set as the sum of the frequency counts of each concept within that set. We then sample 500 instances from both the \commongen and \vocabulary datasets from the top 10\% highest frequency concept sets and 500 each from the bottom 10\% frequency sets. We split the test data this way to evaluate two likely use cases. The lowest 10\% aligns with the case where we have unseen concepts and want to generate something novel; the most frequent 10\% matches the every-day use case where we generate from things we've seen before. This gives us a test set of 2000 total sample, half from \commongen and half from \vocabulary, additionally split into high frequency and low frequency concepts. From the remaining dataset we randomly use 90\% as training and 10\% as validation. 

%\footnote{Please refer to \newcite{lin-2020} for more information related to \commongen.}

%\footnote{\newcite{becker-etal-2021-coco} describes a tool for linking concepts from text such as sentence to ConceptNet

\section{Methods} \label{section:method}

Here, we present our computational approaches for sentence generation using the concept2seq framework. At its base form, our task is to generate a sentence $s$ that consists of sequence of tokens, $s = \{s_1,...,s_m\}$ using a list of concepts $c = \{c_1,...,c_n\}$. Using the standard autoregressive sequence-to-sequence architecture \cite{sutskever2014sequence}  
we model $P_\theta(s\mid c)$ as follows:
\begin{equation} \label{eq:1}
    P_\theta(s\mid c) = \prod_{i}P_\theta(s_i\mid s_1,\dots,s_{i-1},c) 
\end{equation}
\noindent Note, the resulting sentence $s$ should contain relevant vocabulary from the concept set $c$, but is otherwise unconstrained.  We use a pretrained BART model  \cite{lewis-etal-2020-bart} composed of a bidirectional encoder and an autoregressive decoder. In our simplest setup (called \base), the input for BART is a set of concepts $c$ and the output text $s$ is a sentence that contains those concepts. For the \base model, we fine-tune the \texttt{bart-base} model on our training data described above, and then generate sentences based on the test concepts. Note, this is equivalent to the BART based experiment reported in \newcite{lin-2020}.\footnote{Although \newcite{lin-2020} also reported experimental results using other transformer-based LMs such as GPT-2 \cite{radford-2019} and T5 \cite{raffel2020exploring}, we notice BART performs better on several metrics, so we continue to use BART.}

%BART is trained by corrupting text with an arbitrary noise function and training a model to reconstruct the original.

%Another option is to control for the difficulty or complexity of the output. By difficulty we are  here referring the ability level required to understand the generated sentence: in this subtask,

%DG: (if space permits) Educators often refer to the concept of the "zone of proximal development" by providing content which is beyond the skill level of the user, but which can be achieved with the right guidance \cite{doolittle1997vygotsky,mcleod2012zone}. We can better tailor the generated sentences in the LLA to specific users and students based on such theoretical underpinnings.
    
\subsection{\cefr-controlled generation} \label{sec:cefr}
Research in language learning has shown that students' retention of words and texts increases when they encounter increasing diversity in content \cite{adelman2006contextual,frances2020effects}.  Since language learning is affected by many variables \cite{oxford1989variables}, we focus on a specific variable - the CEFR levels - which are an international standard that measures text complexity and has strong correlations with learners' skill sets and language learning ability \cite{papageorgiou2015association}.
%educators are encouraged to focus on three important components which aid the learning process:As stated earlier, one of the motivations of this work is to generate diverse content for different skill levels of users. Thus, we aim to provide the model with additional information that will influence the generation to be aimed at learners with a particular ability level. This allows us to be better tailor the generated sentences to specific users and/or applications. 
To this end, we generate sentences guided towards different skill sets of users by conditioning on the CEFR levels.  These labels are, in increasing order of proficiency: A1, A2, B1, B2, C1, and C2, where A1 denotes a beginner level and C2 denotes high proficiency. Our goal is to be able to start with a list of concepts and generate a sentence at the appropriate proficiency level. We first use a document level CEFR predictor \cite{montgomerie-2022} to predict the CEFR levels for each sentence in the training data. This tagger, which functions by combining lexical, syntactic, and other attested proficiency features, provides a tag from A1 to C2 for each sentence in the training dataset. In turn, we use this predicted CEFR level as control codes to guide sentence generation. 

\begin{figure}[t]
    \centering
    \includegraphics[width=.5\textwidth]{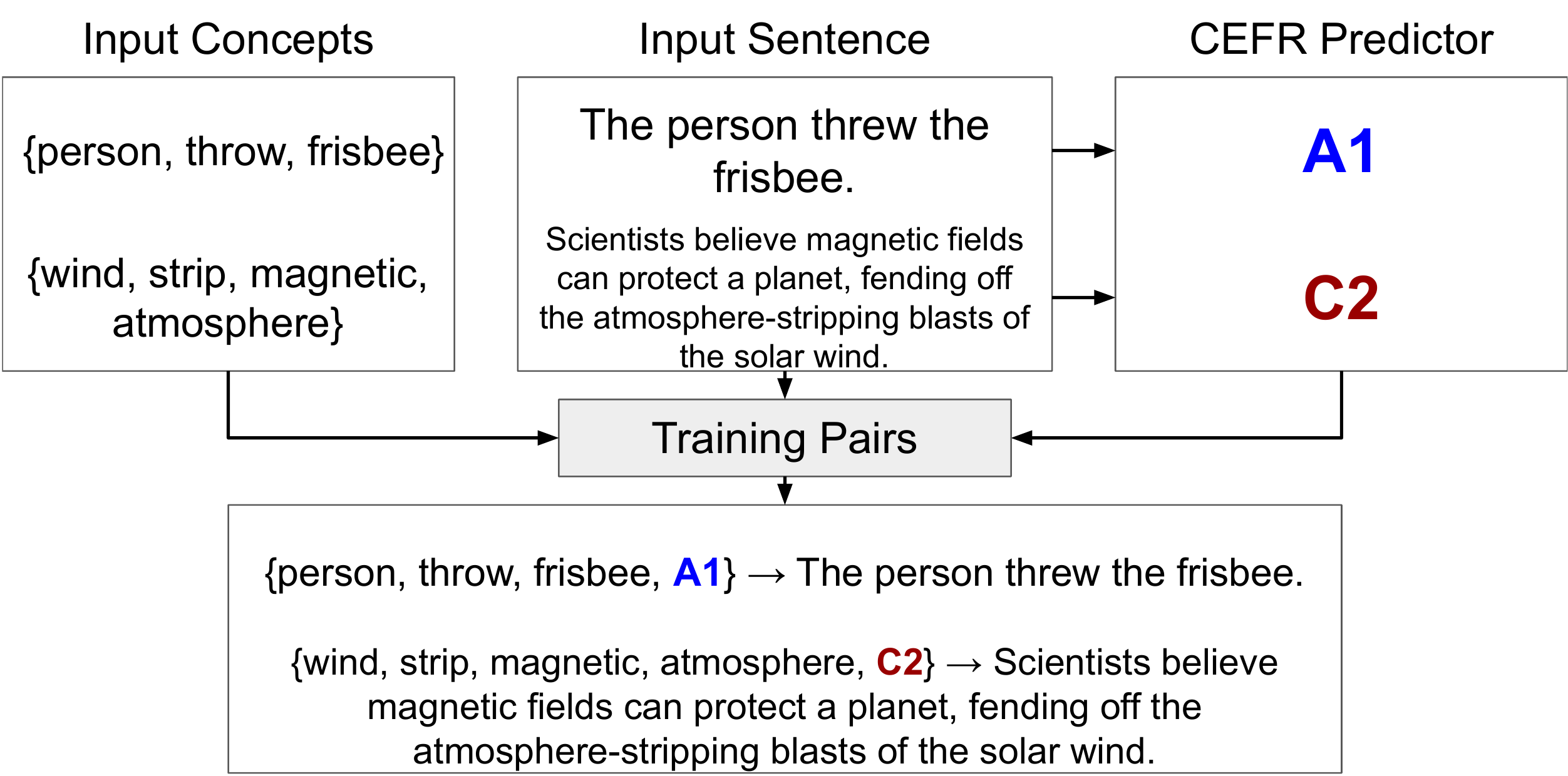}
    \caption{Method for applying CEFR labels to input concepts using the CEFR scorer.}
    \label{fig:cefr}
\end{figure}

Controlled generation models   \cite{kikuchi-etal-2016-controlling,hu2017toward,ficler-goldberg-2017-controlling,tsai2021style} condition on a control code $f$ in addition to the input $c$ to model the distribution of $P_\theta(s\mid c,f)$. 
Similar to Eq.~(\ref{eq:1}), we can write,
\begin{equation} \label{eq:2}
    P_\theta(s\mid c,f) = \prod_i P_{\theta}(s_i\mid s_1,\dots,s_{i-1},c,f) 
\end{equation}
%An overview of this process is shown in Figure \ref{fig:difficulty}.
Text generation conditioned on such control codes, such as sentiment control of movie reviews, style for chatbots, diverse story continuations, question generation etc., have been used effectively in recent research \cite{tu-etal-2019-generating,krause-etal-2021-gedi-generative,roller2020recipes,gao-etal-2022-makes}. We use the same idea for sentence generation by conditioning on the CEFR levels. Figure \ref{fig:cefr} shows the overview of the process.

\subsection{Argument Structure-controlled generation}
    \label{sec:srl}
    
    As the second task, we use the argument structure of the generated sentences as the control code. We determine for any given concept in the input what semantic role that concept should play in the output text. This gives two key advantages: we can ensure the semantic viability of generations in which the concepts make more sense in particular roles, and we can extend the variety of sentences generated by varying the semantic roles of the arguments. Consider the following generation:
    
    \vspace{.5em}
    $\textrm{\{dog, chase, cat\}} \rightarrow {\textrm{(a) the dog chased the cat} \atop \textrm{(b) the cat chased the dog}}$ 
    \vspace{.2em}

    As the concepts are unordered, the model can generate both sentences where (a) the dog is chasing the cat and where (b) the cat is chasing the dog. Stereotypically, we would expect (a), but (b) is a viable reading. By enforcing the semantic roles of the concepts, either with the dog or the cat as the agent, our aim is to be able to more concretely choose which output we'd like to see. This in turn can be utilized to check grammar skills of the users by follow-up questions in an LLA (e.g., which is the agent in the sentence?).
    
\begin{figure}[t]
    \centering
    \includegraphics[width=.5\textwidth]{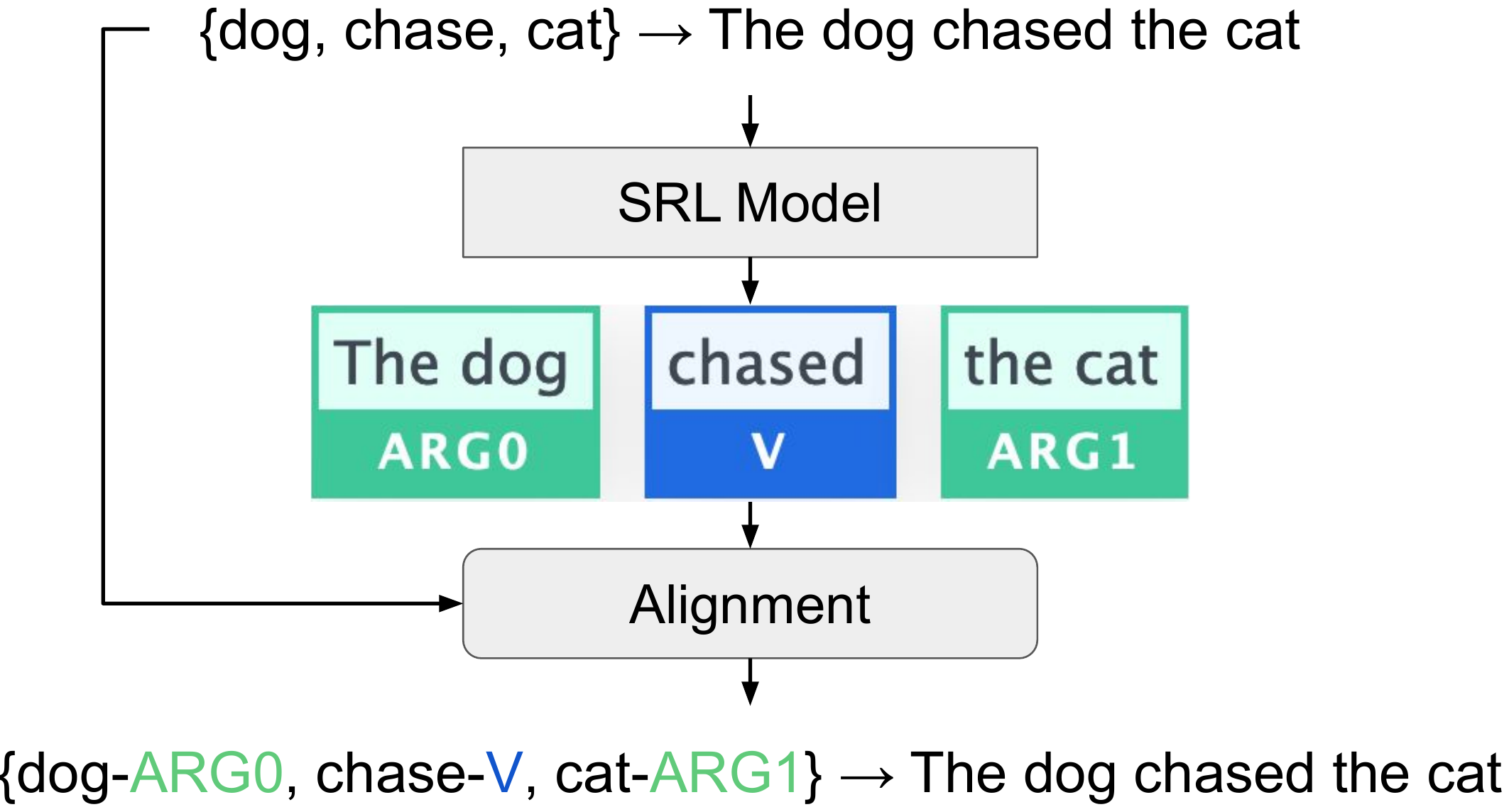}
    \caption{Method for applying SRL parse to the original concepts during training.}
    \label{fig:srl}
    \end{figure}

    To identify the semantic roles, we tag the training dataset with an automatic semantic role labeling system \cite{stanovsky-2018}.\footnote{\url{https://github.com/allenai/allennlp}.} For each verb in the input, the system tags each word in the sentence with the argument it takes in that verb's scope. In order to convert these tags to control codes, we first extract each word in the sentence that matches a lemmatized version of the one of the input concepts. For each of these words, we identify all of the possible roles it can play in a sentence (note that words can take multiple roles, when they are arguments of separate verbs). We then iterate through these options, aligning the possible semantic role labels that word can take to the concepts in the input. This yields a new batch of concepts labeled with semantic role information that serve as the inputs for the given sentence. An overview of this process is shown in Figure \ref{fig:srl}.
    %Similar to the CEFR labels we simply concatenate the semantic roles to the concepts for generation. 
    
\paragraph{Inference} 
For CEFR-controlled generation, we have trained a single model on the full scope of CEFR tagged data: in order to generate sentences at a particularly level, we need to provide that level to the model at inference time. For this, we experiment with the simplest level (A1) and the most advanced level (C2). We add these labels to the concept inputs to generate sentences that should match those levels: these setups are dubbed CEFR-A1 and CEFR-C2.  provided different CEFR levels to generate different sentences.  For argument structure-controlled generation, we run our SRL tagger on the test data and then apply these to the input sources to generate a SRL-controlled output, as was done in training (Figure \ref{fig:srl}).
%We specify the CEFR level (for CEFR-controlled generation) and the semantic roles (for argument structure-controlled generation) to generate specific sentences. 
We generate using top-K sampling ($k=50$) with a maximum length of 64 and a length penalty  of 1.0. 
%To ensure consistency, the first character is capitalized and all sequences end with a single period.
%In turn, such generations can be utilized to check grammar skills of the users by followup questions in an LLA (e.g., which is the agent in the sentence?)
%These two applications, complexity and argument-structure control, give us a new set of input concepts that we use to train BART models. 

%The goal of training these two controlled generations is to give additional control over the output: for any given test sentence, we  apply a CEFR label or SRL to influence the generation. In case of the latter, such generations can be utilized to check grammar skills of the users by follow-up questions in an LLA (e.g., which is the agent in the sentence?)

\subsection{Additional pretraining}
    The above BART model is originally trained with text as both the input and output. Our task is somewhat different, as the input consists of concepts. While these concepts are superficially a set of keywords, this still differs from what the original BART encoders have seen during training. In order to encourage the model to better handle this concept2seq data formulation, we leverage the power of additional pretraining, which has been shown to further improve model performance on new tasks \cite{gururangan-2020}.
 
    We perform additional pretraining using wikipedia. Starting with a dump of wikipedia data,\footnote{\url{https://dumps.wikimedia.org/enwiki/latest/}} we first extract sentences which are then run through our concept extraction pipeline (Section \ref{section:data}). We filter down to 10M random concept-sentence pairs with 2-5 concepts/sentence. These 10M pairs are then used to continue training on top of the pretrained \base model (the \wiki model).\footnote{Full details of the training procedures are in Appendix \ref{app:training}.}

%    \begin{table*}[t]
%        \small
%        \centering
%        \begin{tabular}{l|c|c|c|c|c}
%        \textbf{Model} & \textbf{Perplexity} & \textbf{Coverage (All)} & \textbf{Coverage %(Any)} & \textbf{Length} & \textbf{Diversity}  \\
%        \hline
%        \base & 4.51 & 50.55 & 93.34 & 12.13 & 43.34 \\
%        \hline
%        \srl & 4.53 & 51.40 & 94.10 & 11.98 & 43.51 \\
%        \hline
%        \cefr (A1) & 4.52 & 49.16 & 92.70 & 11.58 & 43.95 \\
%        \cefr (C2) & 4.41 & 39.70 & 74.61 & 15.26 & 37.31 \\
%        \hline
%        \wiki & 4.58 & 51.51 & 94.10 & 12.28 & 42.88 \\        
%        \end{tabular}
%        \caption{Automatic evaluation of generation models. Lower scores indicate better \textbf{Perplexity} and higher \textbf{Diversity}.}
%        \label{tab:baseline}
%    \end{table*}
    
%We also look at a handful of metrics that aren't necessarily tied to the quality of generation, but rather are useful to understand the control of the models: first,

    \begin{table*}[h]
        \centering
        \begin{tabular}{l|c|c|c|c|c}
        \textbf{Model} & \textbf{Perplexity} & \textbf{Coverage (All)} & \textbf{Coverage (Any)} &  \textbf{Length} & \textbf{Diversity}  \\
        \hline
        \base & 4.51 & 50.55 & 93.34 & 12.13 & 43.34 \\
        \hline
        \srl & 4.53 & 51.40 & 94.10 & 11.98 & 43.51 \\
        \hline
        \cefr (A1) & 4.52 & 49.16 & 92.70 & 11.58 & 43.95 \\
        \hspace{2.2em} (C2) & 4.41 & 39.70 & 74.61 & 15.26 & 37.31 \\
        \hline
        \wiki & 4.58 & 51.51 & 94.10 & 12.28 & 42.88 \\        
        \end{tabular}
        \caption{Automatic evaluation of generation models. Lower scores indicate better \textbf{Perplexity} and \textbf{Diversity}.}
        \label{tab:automatic}
    \end{table*}

    \begin{table}
        \small
        \centering
        \begin{tabular}{l|c|c|c|c}
        \textbf{Model} & \textbf{V} & \textbf{ARG0} & \textbf{ARG1}  & \textbf{ARGM} \\
        \hline
        \base & 88.18 & 70.48 & 73.34 & 58.67 \\
        \hline
        \srl & 94.06 & 88.93 & 88.01 & 77.60 \\
     
        \end{tabular}
        \caption{Automatic evaluation of SRL coverage. Scores refer to percentage of times the label occurred with a given concept in the output over the input.}
        \label{tab:srl}
    \end{table}
    
    \begin{table*}
    \centering
    \begin{tabular}{l|c|c|c|c|c|c}
    & \multicolumn{2}{c|}{\textbf{Grammatical}} & \multicolumn{2}{c|}{\textbf{Complex}} & \multicolumn{2}{c}{\textbf{Plausible}} \\
    \cline{2-7}
    \textbf{Model} & \textbf{Common} & \textbf{Rare} & \textbf{Common} & \textbf{Rare} & \textbf{Common} & \textbf{Rare} \\
    \hline
    \base & 3.50 & 3.40 & 2.41 & 2.58 & 3.60 & 3.61 \\
    \hline
    \srl & 3.48 & 3.46 & \textit{2.25} & \textit{2.45} & 3.58 & 3.61 \\
    \hline
    \cefr (A1) & 3.50 & 3.44 & 2.40 & 2.50 & 3.63 & 3.61 \\
    \hspace{2.235em} (C2)  & \textit{3.31} & 3.41 & \textbf{2.67} & \textbf{2.78} & 3.57 & 3.61\\
    \hline
    \wiki & 3.44 & 3.42 & 2.46 & 2.60 & 3.56 & 3.59  \\
    \end{tabular}
    \caption{MTurk evaluation for each generation model, evaluated for grammaticality, complexity, and plausibility. \textit{Italic} scores are significantly below the baseline; \textbf{bold} are significantly above (two-tailed t-test, $p<.01$).}
    \label{tab:human}

\end{table*}

\begin{table*}[t]
\begin{center}
\small
\setlength{\tabcolsep}{4pt}
\begin{tabular}{@{} p{2cm} c p{9cm} ccc@{}}
\toprule
Input Concepts  & Model & Sentence  & Gram & Comp & Plaus

\\ \midrule
\iffalse
\multirow{5}{=}{\{pace, author, contemporary, literature\}} & \base & I'm not interested in literature & 4 & 1.33 & 4 \\ %\cline{2-7} 
& \srl  & I'm not interested in literature. &  4 & 1.33 & 4 \\ %\cline{2-7} 
 & \cefr (A1)  & I'm not very interested in literature. & 4 & 1.67 & 4 \\ %\cline{2-7} 
& \cefr (C2)  & Although many european researchers have studied ancient persian literature in the nineteenth century, the new world is not paying attention to our contemporary literature.  & 3          & 3.33         & 2.67           \\ %\cline{2-7} 
& \wiki   &  He is very interested in literature. & 4 & 2.33 & 4  \\ 

\midrule
\fi
\multirow{5}{=}{\{competition, agree, strange, participate\}} & \base & I agree with you that it's strange to participate in a competition with your dog. & 4 & 3 & 4 \\ %\cline{2-7} 
& \srl  & They agreed to participate in the strange competition. &  4 & 2 & 4 \\ %\cline{2-7} 
 & \cefr (A1)  & I agreed to participate in the strange competition. & 4 & 2.67 & 3.67 \\ %\cline{2-7} 
& \cefr (C2)  & In some cases, it might be safer to think of different words to use if you don't want to participate in a strange competition. & 4 & 3 & 3.67           \\ %\cline{2-7} 
& \wiki   &  In a strange twist of luck, i'd like to participate in a competition. I don't agree with you. & 2 & 3.33 & 2.33  \\

 \midrule
 
\multirow{5}{=}{\{girl, clothes\}} & \base & A girl in black clothes is standing on a ladder and drawing on a wall. & 4 & 2.33 & 4 \\ %\cline{2-7} 
& \srl  & A girl in black clothes is playing soccer. & 4 & 2.33 & 4 \\ %\cline{2-7} 
 & \cefr (A1)  & A girl in black clothes is playing with a toy lawn mower. & 4 & 1.67 & 4 \\ %\cline{2-7} 
& \cefr (C2)  & Young people sit in plastic chairs arranged around a set of stairs in a covered concrete area, wearing swim clothes, resting, and waiting. & 4 & 3 & 4         \\ %\cline{2-7} 
& \wiki   &  A girl in black clothes playing soccer. & 3.33 & 2 & 4 \\

\iffalse
\midrule

\multirow{5}{=}{\{intersection, people, clothing\}} & \base & A group of people in dark clothing are crossing an intersection in a city. & 4 & 2 & 4 \\ %\cline{2-7} 
& \srl  & A group of people in dark clothing are corssing an intersection in a city. & 4 & 2 & 4 \\ %\cline{2-7} 
 & \cefr (A1)  & A group of people in heavy clothing congregate near an intersection. & 3.667 & 2 & 4 \\ %\cline{2-7} 
& \cefr (C2)  &  A motorcycle is standing near a decaying building wall, window, and close-line with several clothing items hanging. & 3.667 & 3.333 & 3.667 \\ %\cline{2-7} 
& \wiki   & A group of people in reflective clothing is crossing an intersection together. & 3.667 & 3 & 3.667 \\

\fi

\bottomrule
\end{tabular}
\caption{\label{table:humananalysis} Examples of generated sentences via different models with their annotated scores.}

\end{center}
\end{table*}

\section{Results} \label{section:results}
    Evaluating natural language generation tasks can be difficult, and some automatic metrics can be problematic \cite{reiter-2018}. To overcome these difficulties, we use metrics specifically tailored to our task, as well as performing manual evaluation to get a concrete understanding of model performance.
    
    %\debanjanin{it is not clear why problematic. due to insufficient references?}
    
    \subsection{Automatic Evaluation}
    Standard metrics, e.g., BLEU \citep{papineni-etal-2002-bleu} or ROUGE-L \citep{lin-2004-rouge} that are often used to evaluate NLG outputs require true reference sentences for evaluation purpose. These methods are insufficient for our approach -- our goal here is to generate sentences containing particular concepts conditioned on specific controls (e.g. CEFR) -- and the resulting outputs do not need to match any particular gold standard. For that reason, we employ the following reference-free metrics for evaluation.
    
    First, \textbf{perplexity} under a language model can indicate the fluency of the text. We report average perplexity per word using the GPT-2-base LM \cite{radford-2019} in the generated sentences. \textbf{Coverage} indicates whether the generated sentences contains the input concepts. We evaluate the percentage of generations that contain lemmas matching the input concepts in two ways: first, the percentage of outputs containing \textbf{any} lemmas matching the input as well as the percentage of those where \textbf{all} of the concepts are found in the output.  We also measure average \textbf{length} of the generated sentences (in number of words). Finally, we measure lexical \textbf{diversity}: for this, we use the average tf-idf score of all non-stopwords in the sentence (learned from a recent Wikipedia dump): higher scores indicate more common words, while lower scores indicate more lexical diversity. Relevant results are shown in Table \ref{tab:automatic}.
%models controlled for difficulty should produce longer sentences when higher difficulty is required.

%as above, models producing more difficult sentences should have higher lexical diversity.
    
%    \subsubsection{Results}
        
    We note a number of observations from automatic metrics: perplexity remains relatively stable across models, indicating they all can produce fluent sentences. \cefr C2 has the lowest perplexity, indicating that BART can produce complex but still fluent sentences. Coverage, length, and diversity remain relatively stable across models as well. One exception is the \cefr C2 model, which has lower coverage (39.70) and higher length (15.26 words per sequence). Since C2 sentences in the training dataset are longer (averaging 25.1 words per sentence, compared to the overall average of 17.0), it is expected that \cefr C2 model  produce longer sentences when higher proficiency is required. Likewise, the \cefr A1 model tends towards shorter sentences (11.58 words per sequence). Finally, the low diversity score of the \cefr C2 model indicate the complex sentences generated by the model have higher lexical diversity.
    
    \paragraph{}{\textbf{SRL Overlap Evaluation}}:
    Note that for the \srl model, we evaluate the test data with a single set of SRL labels generated from the original SRL model. There are many ways to apply SRL labels to a given set of concepts, and we only evaluate against a single reference. 
    
    We use an automatic parser to capture whether the SRL-based inputs are accurately represented in the outputs. For the four most frequent argument types (ARG0, ARG1, ARGM, and V), we evaluate accuracy by comparing its presence in the generated output to its presence in the control codes. We measure the percentage of times the argument type is correctly represented by a concept in the generated sentence over the number of times the instructions indicate it should be. We compare the \srl to the \base model in Table \ref{tab:srl}.

    We might expect a large, pretrained model like BART to automatically generate the concepts into their expected roles, but we can see that the \base model actually fluctuates greatly: for non-verbal arguments, it generates them in the semantic role of the reference sentence less than 75\% of the time. This isn't necessarily a problem, as the system still is generating sentences with the appropriate concepts, but it highlights the usefulness of argument control: using the \srl model, we can generate concepts into specific semantic roles much more consistently, with scores ranging from 77\% to 94\%, thus, improving by a large margin over the \base model.

    %These metrics can serve to give us some basic idea of the models' performance, and to help model and hyperparameter tuning. From here, we then require human evaluation.

    \subsection{Human Evaluation}
    For human evaluation, we aim to capture three essential criteria that are important for test item generation. These are:
    
    \begin{enumeratesquish}
        \item \textbf{Grammaticality}: The generated sentences/phrases should be grammatical, and should follow normal English syntax.
        \item \textbf{Complexity}: The complexity of a given sentence as it relates to end users. %Note, this criteria is useful for the \cefr model the most given we explicitly control that model on different skill levels. 
        \item \textbf{Plausibility}: The generated items should describe semantically plausible scenarios, or they risk confusing or even misinforming the user.
    \end{enumeratesquish}

    We evaluated the generated outputs across these three criteria using Amazon Mechanical Turk (MTurk). We evaluated 800 sentences generated from each model. Three crowd-annotators were employed for each task and were asked to evaluate each sentence on a four-point scale for each criteria. We included many examples in the instructions. Each Human Intelligence Task (HIT) contained ten sentences to judge and we paid \$2 per HIT. We obtain three scores for each of the above criteria for each generated sentence, and take their mean as the final score (Table \ref{tab:human}).

    We observe a number of key take-aways from the human evaluations. First, the rare concept sets are more likely to yield more complex generations, but otherwise they are fairly similar to the common sets: they exhibit similar grammaticality and plausibility scores.  All models score strongly for grammaticality: the \cefr C2 model is lowest, as it is attempting to generate more complex sentences and likely to make more mistakes, but all models average about 3.4. Second, with regard to complexity, the \cefr A1 model scores lower than the \base while the \cefr C2 model scores higher: this is our expected result, as the lower A1 level instruction yields simpler sentences, while the higher level C2 yields more complex sentences. Third, all models perform similarly with regard to plausibility, with every model being within .04 of the baseline. Finally, we see that additional pretraining doesn't improve performance significantly over the baseline: the BART-base model seems perfectly capable of adapting to concept2seq instructions without additional pretraining. 
    
    Table~\ref{table:humananalysis} presents two examples from our models along with average human ratings for all three aspects. In general, \cefr C2 has produced long sentences with high complexity for all examples. Likewise, grammaticality and plausibility scores are almost perfect except one example from \wiki. %We present more examples in the Appendix.  

    %In general, these results show that adding additional control via SRL/CEFR levels doesn't hurt the performance of the model, and allow for some control over the difficulty level. 
    
    %In general, these results facilitate our practical use case: we are able to generate sentences based on certain concepts, and these concepts are grammatical and semantically plausible.
    
    In general, the methods we implemented to allow for additional control (CEFR and SRL) function as expected: we can manipulate the proficiency and argument structure of the generated sentences to a significant degree, allowing us to develop diverse content for users at different levels for LLAs.
    
    \subsection{Performance Time}
    The model was trained on a single nVidia K80 GPU for approximately 158 minutes, at approximately 81 training samples processed per second. We are then able to generate approximately 54 sentences per second at inference time. While this makes the system capable in some regards of generating live learning items, this is not desirable nor is it our use case. There are substantial risks involved in generating items live and presenting them to users, including possible grammatical and semantic disfluencies, unsuitable content, and biases inherent in generation from language models \cite{sheng-2021}. Rather, this system is designed to be run offline, generating a batch of possible learning items that can then be curated by experts. 

\section{Related Work} 
\label{section:re}

The concept2seq generation problem has been investigated in several recent studies. \newcite{lin-2020} released the \commongen dataset and generated sentences using various transformer models, \cite{carlsson-etal-2022-fine} have proposed prompting for generation, and \cite{zhou2020pre} have conducted instruction tuning for generation using concepts. Our work is related to the above and our novelty is that we utilize this framework to generate LLA items. Although we did not experiment with the ordering of concepts similar to \cite{zhao-etal-2022-revisiting}, our \srl based generation in fact implicitly control the order of the concepts by offering specific grammar roles.
%Beside the \commongen data we have also added new sentences (\vocabulary dataset) during training.

%Based on previous work on controllable generation, we generate text that reflects specific characteristics of control variables. 
In prior work on controllable generation, embedding vectors of the control variables were fed into the model to control the output \cite{kikuchi-etal-2016-controlling,fan-etal-2018-controllable}, whereas our approach resembles recent efforts where the control variable is concatenated to the main input \cite{keskar2019ctrl} to control particular style, such as sentiment, style for chatbots, diverse story continuations and argument generation \cite{tu-etal-2019-generating,schiller-etal-2021-aspect,krause-etal-2021-gedi-generative,roller2020recipes}. 

%(formal vs. informal *), sentiment (positive vs. negative *), metaphoricity \cite{stowe-etal-2021-metaphor}, arguments \cite{schiller-etal-2021-aspect}

%We use the same idea for sentence generation by conditioning on the CEFR level $f$ as identified via the tagger.

%or any number of other attributes (*).

\iffalse 

    **Check motivations of relevant work, especially more recent stuff.

    \subsection{Concept2seq generation}
        \begin{itemize}
        \item CommonGen \cite{lin-2020}
        \item Plug n play K2T \cite{pascual-2021}
        \item KG-Bart \cite{lie-2020}
        \item POINTER \cite{zhang-2020}
        \item GDC \cite{khalifa-2021}
        \item Fine-Grained Generation with NRP (ACL 2022, citation coming soon)
    \end{itemize}    

    \subsection{Controlling generation}
    Recent improvements in deep language models have facilitated a bloom in NLG work particularly focused on controlling the outputs. This can be with regard to some particular style (formal vs. informal *), sentiment (positive vs. negative *), metaphoricity (**), or any number of other attributes (*). This control can be implemented by addgin textual control codes to the inputs (*), other methods (***).

    \subsection{Vocabulary/learning item generation?}
        \begin{itemize}
            \item Call the task "concept elaboration" (single/multiple words to a sentence) (Transformer-Based Deep Neural Language Modeling for Construct-Specific Automatic Item Generation, Hommel 2021)
            \item Difficulty control for distractors \cite{yeung-2019}
        \end{itemize}

\fi
\section{Conclusion and Future Work} \label{section:conclusion}
We proposed a type-controlled sentence generation framework for LLAs. We generate sentences (a) conditioned on the CEFR levels to provide content for users/students who belong to different skill sets (e.g., beginner or proficient in English), and (b) conditioned with specific argument structures for grammar. In automatic evaluation, the \srl model shows better coverage of input concepts than \base, whereas human evaluation demonstrates high grammatically scores (3.4 and above) for all the models as well as high complexity for the \cefr C2 model that was designed to generate complex sentences for proficient users. In future, we want to continue a couple of error analyses on the input as well as on the generated sentences. Having taken into account that input data is pre-processed in several ways (e.g., concept extraction \cite{becker-etal-2021-coco} and analysis of semantic roles \cite{stanovsky-2018}), we want to select a small subset of data to determine whether such extraction has any error. Likewise, we also want to employ expert content developers to analyze the results of the CEFR predictor. Finally, we plan to employ additional controls such as word senses to guide context specific generations.

\section*{Acknowledgments}
Thanks to Casey Medlock Paul and Kristen Herrick for suggesting reference materials on learning science, as well as Swapna Somasundaran for helpful comments.
\section{Ethical Considerations} \label{section:ethics}

We leverage the freely available \commongen dataset for model training. Though we have not exhaustively checked the dataset, given \commongen is based on a variety of caption datasets, we consider them relatively safe and do not find any objectionable content. Likewise, we create another dataset, \vocabulary, which is based on standard narratives and sentence databases that are used in many recent work. Training is done using large pretrained models that have been shown to have bias. Although the generated content do not appear biased, they may hallucinate content, which is a common problem for neural generation models. In future work, we plan to analyze and identify hallucinations from the generations, and assess possible bias issues within these generations.

Finally, we obtained institutional review board permission to conduct MTurk based evaluations to collect judgments from crowd workers regarding the quality of the sentences.

\bibliography{anthology,custom}
\bibliographystyle{acl_natbib}

\appendix
\section{Appendix} \label{section:app}

\subsection{Vocabulary items}
The following fourteen words in Table \ref{tab:vocabcefr} (see their associated CEFR levels) were suggested by the learning scientists/content developers of the LLA we are involved with. 

\begin{table}[h!]
        \centering
        \begin{tabular}{c|c}
        \textbf{Word} & \textbf{CEFR}   \\
        \hline
Clothes & A1 \\
Famous & A1 \\
Electric & A2\\
Return & A2\\
Lose & B1\\
Delicious & B1\\
Entertainment & B1\\
Literature & B1\\
Atmosphere & B2\\
Participate & B2\\
Awkward & B2\\
Solar & B2\\
Devote  & B2\\
Caution  & C1\\
\end{tabular}
        \caption{Words that are used to generate the \vocabulary dataset (with their CEFR levels).}
        \label{tab:vocabcefr}
\end{table}

\subsection{Concept Extraction}
We extracted concepts using the concept extractor tool CoCo-Ex \cite{becker-etal-2021-coco}. The tool first parse sentences using standard parsers and then match tokens as found in ConceptNet knowledge base. The resulted concepts are categorized to their parts-of-speech. For our work we use nouns, verbs, and adjective tokens.

\subsection{Model Training}
\label{app:training}
We train our models using the HuggingFace platform \cite{wolf-2020}. 
We use the \texttt{bart-base} model as the initial checkpoint from the HuggingFace repository \cite{wolf-2020}. Each model is trained for 5 epochs with a batch size of 32 and a learning rate is 5e-5, as these parameters yielded the best performance on the validation set. For the CEFR generation, the label is added as an additional concept. For the SRL-based generation, the labels are concatenated to individual concepts. For the additional pretraining with the Wikipedia data, we ran pretraining for 3 epochs. All experiments were conducted using NVIDIA K-80 GPUs.

\subsection{MTurk Experiments}

\begin{figure*}[t]
    \centering
    \includegraphics[width=\textwidth]{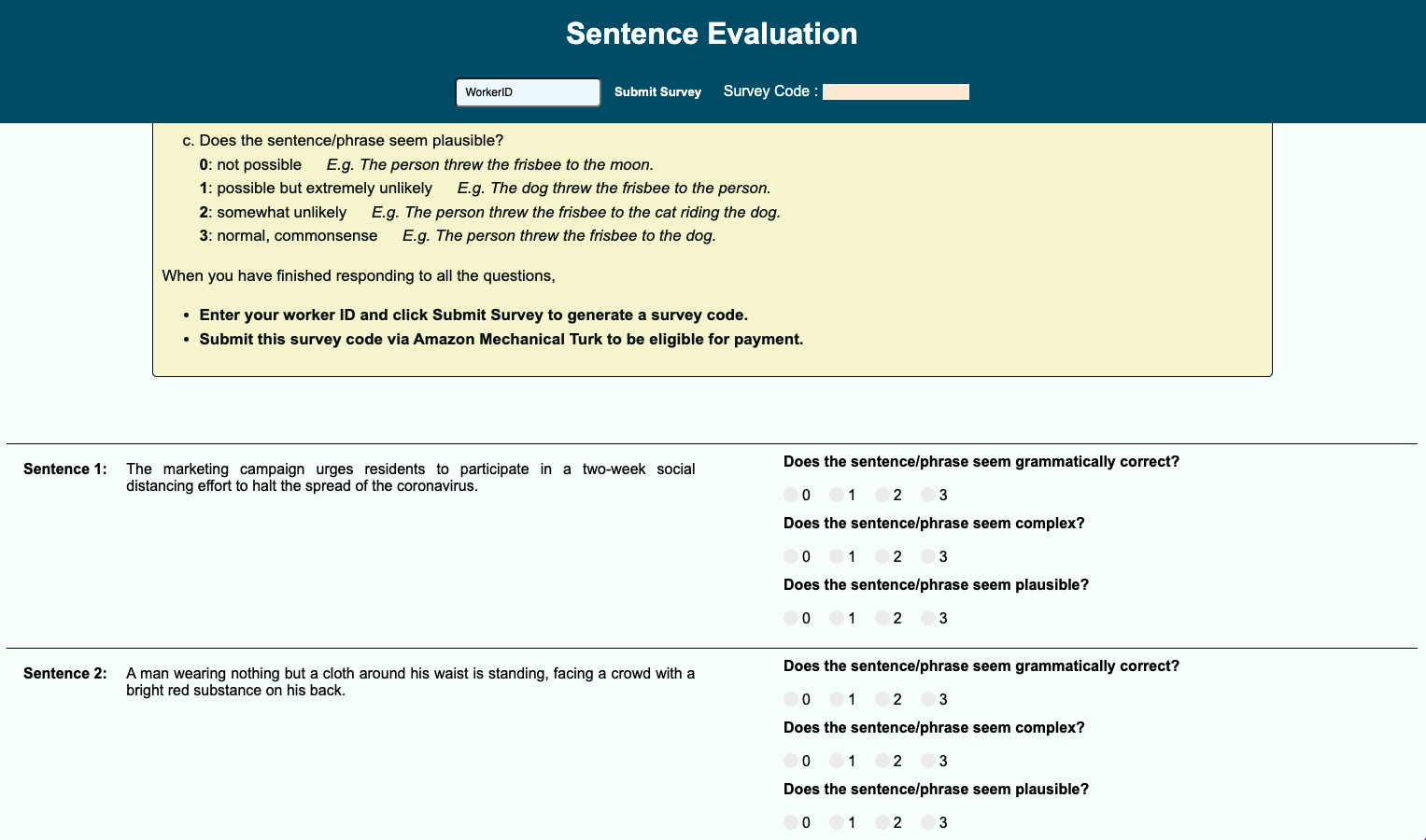}
    \caption{The survey interface where Turkers can read detailed instructions and complete the evaluation of ten sentences.}
    \label{fig:interface}
\end{figure*}

In order to collect human evaluation for generated sentences, we deployed our data collection pipeline using AWS infrastructures. After reading and confirming the consent page, Turkers are directed to the survey interface where detailed instructions and survey questions are presented (shown in Figure \ref{fig:interface}). Turkers must complete all questions to be able to submit. We initially began evaluation using 5 samples per HIT, but extended this to 10 as the time necessary for annotators to complete a HIT was extremely short. A ``survey code'' is returned to the Turker as the submission is successful, and with the code the Turker can submit the HIT and to be qualified for payment. All evaluation results are saved into a DynamoDB table through API Gateway and Lambda.

To ensure the quality of evaluation, we recruited Turkers with more than 10,000 approved HITs and the approval rate or 97\% and above. One task is required to be completed within 60 minutes, and most Turkers were able to submit within 10 to 15 minutes.

\subsection{Human Evaluation}
Table \ref{table:humananalysisapp} presents more examples of MTurk annotations.
\begin{table*}[t]
\begin{center}
\small
\setlength{\tabcolsep}{4pt}
\begin{tabular}{@{} p{2cm} c p{9cm} ccc@{}}
\toprule
Input Concepts  & Model & Sentence  & Gram & Comp & Plaus

\\ \midrule
\multirow{5}{=}{\{pace, author, contemporary, literature\}} & \base & I'm not interested in literature & 4 & 1.33 & 4 \\ %\cline{2-7} 
& \srl  & I'm not interested in literature. &  4 & 1.33 & 4 \\ %\cline{2-7} 
 & \cefr (A1)  & I'm not very interested in literature. & 4 & 1.67 & 4 \\ %\cline{2-7} 
& \cefr (C2)  & Although many european researchers have studied ancient persian literature in the nineteenth century, the new world is not paying attention to our contemporary literature.  & 3          & 3.33         & 2.67           \\ %\cline{2-7} 
& \wiki   &  He is very interested in literature. & 4 & 2.33 & 4  \\ 

\midrule

\multirow{5}{=}{\{competition, agree, strange, participate\}} & \base & I agree with you that it's strange to participate in a competition with your dog. & 4 & 3 & 4 \\ %\cline{2-7} 
& \srl  & They agreed to participate in the strange competition. &  4 & 2 & 4 \\ %\cline{2-7} 
 & \cefr (A1)  & I agreed to participate in the strange competition. & 4 & 2.67 & 3.67 \\ %\cline{2-7} 
& \cefr (C2)  & In some cases, it might be safer to think of different words to use if you don't want to participate in a strange competition. & 4 & 3 & 3.67           \\ %\cline{2-7} 
& \wiki   &  In a strange twist of luck, i'd like to participate in a competition. I don't agree with you. & 2 & 3.33 & 2.33  \\

 \midrule
 
\multirow{5}{=}{\{girl, clothes\}} & \base & A girl in black clothes is standing on a ladder and drawing on a wall. & 4 & 2.33 & 4 \\ %\cline{2-7} 
& \srl  & A girl in black clothes is playing soccer. & 4 & 2.33 & 4 \\ %\cline{2-7} 
 & \cefr (A1)  & A girl in black clothes is playing with a toy lawn mower. & 4 & 1.67 & 4 \\ %\cline{2-7} 
& \cefr (C2)  & Young people sit in plastic chairs arranged around a set of stairs in a covered concrete area, wearing swim clothes, resting, and waiting. & 4 & 3 & 4         \\ %\cline{2-7} 
& \wiki   &  A girl in black clothes playing soccer. & 3.33 & 2 & 4 \\

\midrule

\multirow{5}{=}{\{intersection, people, clothing\}} & \base & A group of people in dark clothing are crossing an intersection in a city. & 4 & 2 & 4 \\ %\cline{2-7} 
& \srl  & A group of people in dark clothing are corssing an intersection in a city. & 4 & 2 & 4 \\ %\cline{2-7} 
 & \cefr (A1)  & A group of people in heavy clothing congregate near an intersection. & 3.667 & 2 & 4 \\ %\cline{2-7} 
& \cefr (C2)  &  A motorcycle is standing near a decaying building wall, window, and close-line with several clothing items hanging. & 3.667 & 3.333 & 3.667 \\ %\cline{2-7} 
& \wiki   & A group of people in reflective clothing is crossing an intersection together. & 3.667 & 3 & 3.667 \\

\midrule

\multirow{5}{=}{\{stand, rock, water, body\}} & \base & The man stands on the rocks near the body of water. & 3.333 & 2.667 & 4 \\ %\cline{2-7} 
& \srl  & A man stands on a rock near a body of water. & 4 & 2.333 & 4 \\ %\cline{2-7} 
 & \cefr (A1)  & A body of water standing on rocks. & 3.333 & 2 & 2 \\ %\cline{2-7} 
& \cefr (C2)  & Young man standing on rocks near body of water in the mountains. & 3.333 & 2.667 & 4 \\ %\cline{2-7} 
& \wiki & A man stands on a rock near a body of water. & 4 & 2.333 & 4 \\

\midrule
\multirow{5}{=}{\{sand, beach, group, sit, large, clothing\}} & \base &  A man in black clothing and a black hat is walking in front of a yellow wall. & 4 & 2.667 & 4 \\ %\cline{2-7} 
& \srl  & A man in black clothing walking down a sidewalk. & 3 & 2 & 4 \\ %\cline{2-7} 
 & \cefr (A1)  & A man in very torn, dirty clothing is walking down a city street in front of a well dressed woman. & 4 & 3.333 & 4 \\ %\cline{2-7} 
& \cefr (C2)  & A young man in colorful clothing and a helmet is walking down the street in front of a well dressed woman and young girl. & 4 & 2.667 & 4 \\ %\cline{2-7} 
& \wiki & A woman wearing heavy makeup and a pink jacket is walking past a clothing shop at night. & 4 & 3 & 4 \\

\midrule
\multirow{5}{=}{\{kosovo, game, participate\}} & \base & In kosovo, they participate in a game of hopscotch. & 3 & 2.333 & 4 \\ %\cline{2-7} 
& \srl  & Kosovo will participate in the olympic games. & 3.667 & 1.667 & 3.667 \\ %\cline{2-7} 
 & \cefr (A1)  & In 2006, the olympic games were cancelled after kosovo decided to participate in the oclolympic game. & 3.667 & 2.667 & 1 \\ %\cline{2-7} 
& \cefr (C2)  & The olympic athlete participates in his first match as a professional athlete in the olympics. & 3.333 & 2 & 4 \\ %\cline{2-7} 
& \wiki & I will participate in the kosovo olympic games. & 3.667 & 1.667 & 3.667 \\

\bottomrule
\end{tabular}
\caption{\label{table:humananalysisapp} Examples of generated sentences via different models with their annotated scores.}

\end{center}
\end{table*}

\end{document}